\documentclass[letterpaper, 10 pt, conference]{ieeeconf}
\IEEEoverridecommandlockouts
\usepackage{cite}
\usepackage{amsmath,amssymb,amsfonts}
\usepackage{algorithmic}
\usepackage{graphicx}
\usepackage{textcomp}
\usepackage{xcolor}
\usepackage{comment}
\usepackage{makecell}
\usepackage[colorlinks=true,
allcolors=blue]{hyperref}
\def\BibTeX{{\rm B\kern-.05em{\sc i\kern-.025em b}\kern-.08em
    T\kern-.1667em\lower.7ex\hbox{E}\kern-.125emX}}
\begin{document}

\title{
Effects of fiber number and density on fiber jamming: \\
Towards follow-the-leader deployment of a continuum robot
}

\author{Chen~Qian, Tangyou~Liu, and~Liao~Wu,~\IEEEmembership{Member, IEEE}
\thanks{This work was supported in part by the Australian Research Council under Grant DP210100879, in part by Heart Foundation under Vanguard Grant 106988, and in part by UNSW Engineering under GROW Grant PS69063.}
\thanks{C. Qian, T. Liu, and L. Wu are with the School of Mechanical and Manufacturing Engineering, University of New South Wales, Sydney, Australia. {\tt\small dr.liao.wu@ieee.org}}
}

\maketitle

\begin{abstract}
Fiber jamming modules (FJMs) offer flexibility and quick stiffness variation, making them suitable for follow-the-leader (FTL) motions in continuum robots, which is ideal for minimally invasive surgery (MIS).
However, their potential has not been fully exploited, particularly in designing and manufacturing small-sized FJMs with high stiffness variation.
Although existing research has focused on factors like fiber materials and geometry to maximize stiffness variation, the results often do not apply to FJMs for MIS due to size constraints.
Meanwhile, other factors such as fiber number and packing density, less significant to large FJMs but critical to small-sized FJMs, have received insufficient investigation regarding their impact on the stiffness variation for FTL deployment.
In this paper, we design and fabricate FJMs with a diameter of 4mm. Through theoretical and experimental analysis, we find that fiber number and packing density significantly affect both absolute stiffness and stiffness variation. Our experiments confirm the feasibility of using FJMs in a medical FTL robot design. The optimal configuration is a 4mm FJM with 0.4mm fibers at a 56\% packing density, achieving up to 3400\% stiffness variation.
A video demonstration of a prototype robot using the suggested parameters for achieving FTL motions can be found at \href{https://youtu.be/7pI5U0z7kcE}{https://youtu.be/7pI5U0z7kcE}.

\color{black}

\end{abstract}


\section{Introduction}
\begin{figure*}[t]
    \centering
    \includegraphics[width=1\textwidth]{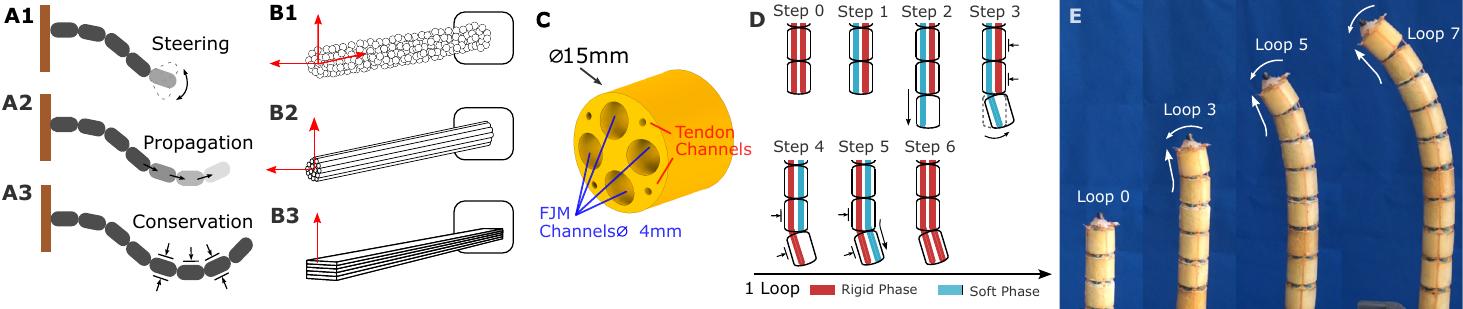}
    \caption{\textbf{(A)} Three essential functions to achieve FTL: \textbf{(A1)} Steer the leader or end-effector. \textbf{(A2)} Propagate along the desired trajectory. \textbf{(A3)} Conserve the trajectory's shape. \textbf{(B)} Three categories of jamming mechanism: \textbf{(B1)} granular jamming, \textbf{(B2)} fiber jamming and \textbf{(B3)} layer jamming. \textbf{(C)} Design of continuum robot that contains tendons and FJMs to achieve FTL. \textbf{(D)} Prototype illustrating FTL motion by changing the stiffness of two FJMs: Step 0: All FJMs are jammed. Step 1: Unjam one FJM. Step 2: Propagate the unjammed FJM. Step 3: Pull tendons to steer the tip. Step 4: Jam the first FJM and unjam the second FJM. Step 5: Propagate the second FJM to the first's position. Step 6: Jam the second FJM. The robot is now ready for the next loop. \textbf{(E)} Photos of the prototype demonstrating FTL motion with four tubes.}
    \label{FTL}
\end{figure*}

With the increasing requirement in complexity and precision for robots in the minimal invasive surgery (MIS) application, Follow-The-Leader (FTL) becomes a critical strategy of instrument navigation that ensures precise manipulation and optimal positioning in confined surgical spaces, such as transoral surgery and transanal surgery \cite{culmone2021follow,li2017kinematic}.
FTL motion refers to the behavior where a continuum robot follows the path of its tip. As illustrated in Fig.~\ref{FTL}A, continuum robots must perform three functions to achieve FTL: steering, propagation, and conservation \cite{culmone2021follow,choset1999follow}.

One popular approach to achieve these functions is introducing variable stiffness modules \cite{zhong2020novel}. Such a design usually consists of components that can transition between soft and rigid phases \cite{shang2022review,manti2016stiffening}. 
By steering and propagating the robot in the soft phase and conserving the path's shape in the rigid phase, the stiffness variation mechanism can potentially enable FTL \cite{degani2006highly}.

Compared to other methodologies to achieve stiffness variation, the jamming mechanism shows advantages in responding time \cite{wang2018development,xing2020structure}, shape adaptability \cite{li2019flexible}, and structural simplicity \cite{yang2020geometric,wang2022design}, which show potentials in FTL applications \cite{culmone2021follow}. A jamming module usually comprises a soft membrane and the filling inside. 
According to the geometry of the filling, jamming modules can be categorized into three types: granular (Fig.~\ref{FTL}B1), fiber (Fig.~\ref{FTL}B2), and layer (Fig.~\ref{FTL}B3) jamming.
When no pressure is applied, these mechanisms exhibit soft properties, whereas when high pressure is applied (usually by using a vacuum to jam the filling), they become rigid \cite{fitzgerald2020review}.

Jamming modules have been extensively investigated in surgical applications and demonstrated promising results in actuators \cite{ranzani2015bioinspired,jiang2013stiffness,brancadoro2019toward}. 
However, despite the distinct advantages of jamming in terms of stiffness changes, there is currently little research on jamming in FTL applications.

To fill this gap, we are designing a new jamming-based robot (Fig.~\ref{FTL}C) to perform FTL mechanisms in surgical applications such as transanal \cite{atallah2010transanal,yu2016development,wu2017development} and transoral surgery \cite{sheth2014vitro}.
The robot design consists of four tendons for tip orientation and four jamming modules to maintain its shape during movement (Fig.~\ref{FTL}D).   
Fiber jamming is chosen for its ability to prevent shape change longitudinally while allowing it in other directions, making it ideal for continuum robots.
Our prototype has an outer diameter of 15mm, suitable for surgery \cite{manfredi2021endorobots,atallah2010transanal,sheth2014vitro}, with a 4mm FJM membrane inner diameter (Fig.~\ref{FTL}C). 
However, building a continuum robot with such micro-sized FJMs without compromising the stiffness variation is challenging, and many factors may affect the stiffness variation, including fiber and membrane materials, jamming geometry, fiber number, packing density, etc. \cite{jadhav2022variable,jiang2014robotic,aktacs2021modeling}.
Therefore, an overall consideration of these factors is crucial to fabricating the micro-sized FJM, which is the focus of this paper.


Previous studies have focused on materials and geometry to improve stiffness variation. 
Margherita \textit{et al.} tested six different fiber materials and found wax cotton provided the highest stiffness variation ratio, but nylon FJM was the stiffest overall \cite{brancadoro2018preliminary}. 
Saurabh \textit{et al.} conducted tests on the abrasive cords in 2022 and found it showed a better performance in stiffness \cite{jadhav2022variable}. 
By changing the membrane materials, Jiang \textit{et al.} \cite{jiang2014robotic} obtained a higher stiffness with a polythene sheet.
%
Additionally, Nikolaos \textit{et al.} \cite{vasios2019numerical} and Buse~\textit{et al.} \cite{aktacs2021modeling} generated the fiber models of FJMs to predict the force-deflection behavior of fibers when pressurized, separately. 
The fiber bundle arrangement was also proved to impact the stiffness \cite{brancadoro2020fiber}. 

Unfortunately, previous research is not applicable to smaller FJMs. For example, the fiber material in earlier studies had diameters exceeding 1mm, which is unsuitable for a 4mm diameter FJM. Additionally, the fiber arrangement and geometry are not feasible for a micro-sized FJM. Furthermore, previous designs mainly focused on high payload or interlocking capacity \cite{ranzani2015bioinspired,jiang2013stiffness,brancadoro2019toward}, but FJMs for FTL motion need high stiffness variation and adequate flexibility when unjammed. Thus, downsizing FJMs for MIS requires careful configuration.



For the above reasons, this paper contributes to fiber jamming in the following two aspects:

\begin{enumerate}
    \item We conduct a fundamental study on the factors not thoroughly examined in previous research, including fiber number, fiber length, the ratio of membrane size to fiber size, and packing density. We find that fiber number and packing density significantly impact results due to friction in micro-sized FJMs, often neglected in earlier studies.
    \item We develop and manufacture micro-sized FJMs with a 4mm diameter. When previous research factors are not applicable, we optimize the configuration for absolute stiffness and stiffness variation, enhancing the feasibility of FTL motion in our robot design.
\end{enumerate}

\section{Theoretical Study}
In this section, we propose formulas to derive and summarize how fiber number and packing density will affect the absolute stiffness of jammed and unjammed FJMs and the relevant stiffness variation ratio.

\subsection{Deflection}
\begin{figure}[t]
    \centering
    \includegraphics[width=0.5\textwidth]{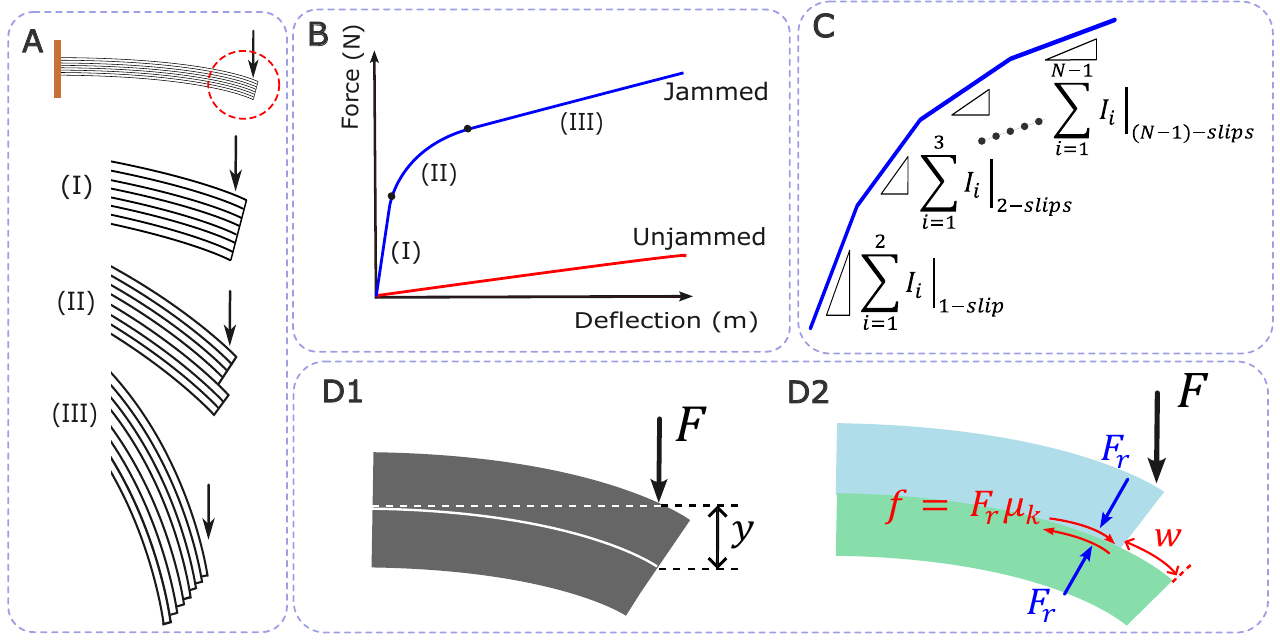}
    \caption{\textbf{(A)} Deflection of an FJM under vertical force in three phases: \textbf{(I)} ``pre-slip" phase: No slipping between fibers, FJM behaves as one single beam. \textbf{(II)} ``transition regime" phase: shear stress exceeds static friction, fibers slip vertically in layers, horizontal fibers remain unslipped. \textbf{(III)} ``full-slip" phase: All fiber layers slip, making jammed and unjammed FJM stiffness similar. \textbf{(B)} Predicted stiffness curves (force vs deflection) of jammed and unjammed FJMs. \textbf{(C)} illustrative stiffness change due to change of moment of the inertia in the ``transition regime" phase. \textbf{(D)} Free body diagram of the fiber bundle when not slipped \textbf{(D1)} and slipped \textbf{(D2}).}
    \label{ST}
\end{figure}

When a non-longitudinal force is applied to an FJM, the fiber bundle starts to bend in the direction of the force.  Shear stress between fibers continues to increase as deflection increases. As illustrated in Fig.~\ref{ST}A, the fibers are cohesive and behave as one beam until the shear stress exceeds the friction force in one of the many contact surfaces of fibers \cite{xiao2016influence}. When the shear stress is high enough to cause all possible slipping between fibers, the stiffness of the FJM reduces to the minimum. Such phenomenon can be summarised into three phases: ``pre-slip", ``transition regime" or ``full-slip" phases \cite{narang2018mechanically}. It is noted that an unjammed FJM usually skips ``pre-slip" and ``transition regime" phases when deflecting because the frictional resistance between fibers is much weaker than deflection shear when no pressure is applied. 

Due to the mechanical similarity, an FJM in either phase can be analyzed as one or multiple cantilever beam(s). When an external force is applied to cantilever beams, the deflection of the FJM can then be expressed as:
\begin{equation}
\begin{split}
    y=\frac{\partial U}{\partial F} & =\frac{\partial}{\partial F}(\int_0^l \frac {M^2(x)}{2EI}dx - \sum \limits^N f\cdot d ) \\
    & = C_1\frac{FL^3}{EI}-\frac{\partial}{\partial F}(\sum \limits^N f\cdot d) ,
    \label{eq1}
\end{split}
\end{equation}
where $C_1$ is a constant value depending on the location and distribution of the external force on the beam, $U$ is the elastic bending energy, $y$ is the deflection distance of the beam along the direction of load, $E$ is the Young’s modulus of the fibers, $I$ is the total moment of inertia of the transverse plane of fibers, $M(x)$ is the bending moment at distance of $x$ on the beam, $N$ is the fiber number, $f$ is the average friction work in fibers, and $d$ is the travel distance of friction force.

It is reasonable to assume the travel distance $d$ is proportional to the deflection $y$. Similarly, the friction force $f$ is expected to be proportional to the external Force $F$. 
\begin{equation}
    d \varpropto y \varpropto \frac{FL^3}{EI} \:\:\:and\:\:\: f \varpropto F .
\label{d&f}
\end{equation}

\subsection{Moment of inertia}
Given that the number of fibers in FJM is N and the radius of fiber is $r$, the total moment of inertia of unjammed FJM $I_{uj}$ is theoretically equal to:
\begin{equation}
    \displaystyle I_{uj}=N\cdot\frac{\pi}{4}\:r^4 \, .
\label{I_uj}
\end{equation}

\begin{figure}[t]
    \centering
    \includegraphics[width=0.4\textwidth]{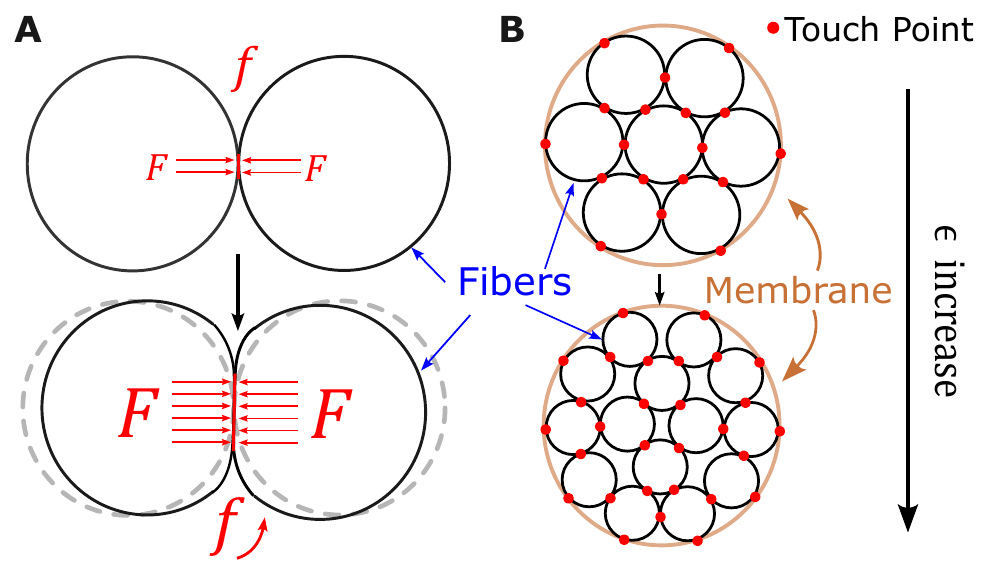}
    \caption{Illustrative factors that increase the coefficient $\epsilon$, which leads to greater friction: \textbf{(A)} The increasing friction force due to higher packing density. \textbf{(B)} The increasing number of contact areas due to more fibers filled in.}
    \label{K-factor}
\end{figure}
In comparison, the calculation of $I_{j}$ is complicated: the cross-sectional area of a jammed FJM is hard to predict. An approximating solution is to consider the cross-section of a circle and $I_{j}$ is equal to:
\begin{equation}
    \displaystyle I_{j}=\frac{\pi}{4}\:R^4 \, .
\label{I_j}
\end{equation}

However, there is no unique formula to determine the radius of a big circle $R$ when N small circles are packed inside with a radius of $r$, as described in the ``circle packing problem" \cite{huang2011global}. Therefore, we determine the overall radius $R$ for each testing scenario and then calculate the number of fiber N accordingly. It is detailed in the later section.

\begin{figure*}[!h]
    \centering
    \includegraphics[width=1\textwidth]{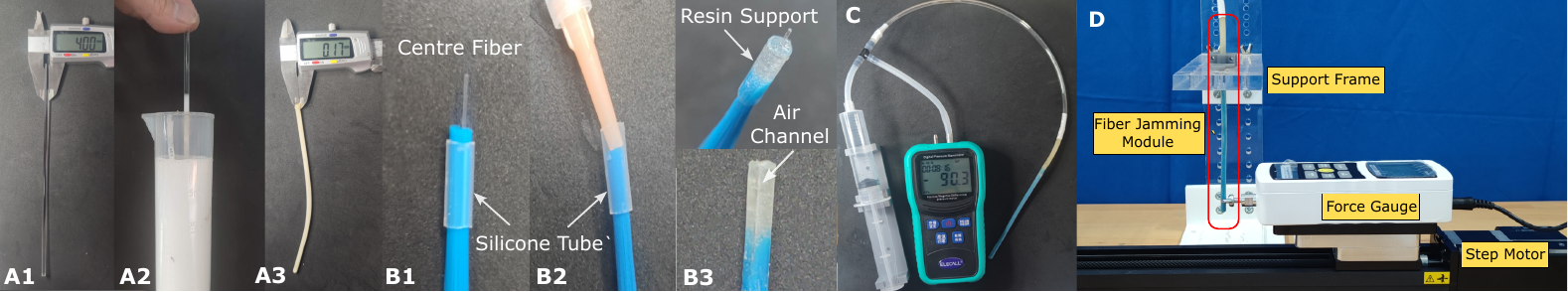}
    \caption{Manufacturing procedure of FJM: \textbf{(A1)} Acrylic rod in 4mm diameter. \textbf{(A2)} Dipping the rod into a latex container. \textbf{(A3)} Membrane thickness is measured at 0.16mm-0.18mm. \textbf{(B1)} Fiber bundle bounded by the molding tube. \textbf{(B2)} Injecting epoxy resin. \textbf{(B3)} The air hole for vacuuming is generated after removing the centre fiber. \textbf{(C)} Vacuum device is composed of one 50ml syringe, a digital manometer and a 3d-printed frame. -90.3kpa is measured when the syringe is forced to vacuum by the frame. \textbf{(D)} Experimental setup for stiffness test.}
    \label{Fabricate}
\end{figure*}

\subsection{Stiffness Variation}

The maximum defection in equation (\ref{eq1}) occurs when the external work is completely converted into the elastic potential energy of the fibers. 
If some of the work is dissipated as heat due to friction, the defection decreases, and the stiffness hence increases. For a jammed FJM, such dissipation is minor because the travel distance $d$ of the friction force is neglectable in the ``pre-slip" phase (Fig.~\ref{ST}D1). However, $d$ will increase as the external increase in ``full-slip" phase and hence increase the stiffness of an unjammed FJM (Fig.~\ref{ST}D2).
To simplify the equations, we introduce an affecting coefficient $\epsilon$ to represent the effect of friction work in the equation (\ref{eq1}) after differential. The stiffness of jammed and unjammed FJMs can then be expressed as:
\begin{equation}
	\frac{F}{y}\Big|_{x=L} =
	\begin{cases}
		\displaystyle \frac{1}{C_1}\frac{E I_{j}}{L^3} & \text{jammed} \\
		\\
		\displaystyle \frac{1}{C_1}\frac{E I_{uj}}{L^3}\frac{1}{1-\epsilon} & \text{unjammed}
	\end{cases} \, ,
	\label{stiffness}
\end{equation}

After merging equations (\ref{I_uj}), (\ref{I_j}) and (\ref{stiffness}), the stiffness variation ratio $\zeta$ of an FJM between jammed and unjammed phases is equal to:
\begin{equation}
    \zeta = \frac{R^4}{N r^4}(1-\epsilon)\\
    \:\:\:and\:\:\: \epsilon \varpropto f,N .
\label{stiffness_ratio}
\end{equation}

As illustrated in Fig.~\ref{K-factor}, $\epsilon$ positively correlates with packing density and fiber number. This conclusion will be verified through practical experiments outlined in the following section.

\section{Experimental Study}

To verify the theoretical deduction in the above section, we fabricated various FJMs for stiffness tests. We normalized fabrication and testing processes to minimize the impact of other irrelevant variables.

\subsection{Fabrication}
FJM comprises a tube-shaped membrane, a fiber bundle, and a vacuuming device. This section provides the fabrication details of each part.

\subsubsection{Membrane}
Latex is selected as the membrane material due to its high stretchiness and barrier properties \cite{jiang2014robotic}. It is also a safe material for medical applications. Membranes are manufactured with the following steps (illustrated in Fig.~\ref{Fabricate}A):
\begin{enumerate}
  \item Prepare acrylic rods with rounded heads for molding. The diameters of these rods are 4mm.
  \item Dip the rods into a latex container with the desired length (effective length + 30mm).
  \item Shake off excess latex immediately after the rod is removed from the container. This procedure helps the latex membranes maintain similar thickness.
  \item After the latex is cured, repeat steps 2 and 3 one more time for sufficient thickness of membrane. 
  \item Peel off the membrane from the rod and check its air-tightness. An approximate thickness of 0.16 mm-0.18 mm is achieved for each membrane. 
\end{enumerate}

\subsubsection{Fiber bundle}
Fiber size can be determined by the fiber number and packing density. However, considering that the fiber size cannot be customized for each test data set, we choose the fibers with diameters of 0.3mm, 0.4mm, 0.5mm, and 0.6mm to ensure a reasonable fiber number in each FJM. Table~\ref{exp_dataset} shows the number of fibers $N$ in each FJM, which is determined by the required packing density. The effective length of fibers is 100mm for all tested FJMs.

Nylon NA66 industrial brush fibers are then selected for fiber materials that have been tested comparatively stiff \cite{brancadoro2018preliminary}. In addition, although their surface finish is not ideal for jamming mechanism, such fibers are standardized production which will promote experimental results closer to theoretical data.

\begin{table}[t]
\centering
\caption{Summary of fiber number for each testing configuration} \label{exp_dataset}
\begin{tabular}{cccc}
\hline
Overall Diameter ($2R$) & Fiber         & No. of    &Packing \\ 
(\% of Membrane Diameter) & Diameter ($2r$)    & Fiber     &Density\\ \hline
 & 0.3mm & 128 & 72.0\% \\ 
3.8mm (95\%)  & 0.4mm & 72  & 71.9\% \\ 
  & 0.5mm & 46  & 72.0\% \\ 
  & 0.6mm & 32  & 72.0\% \\ \hline 
& 0.3mm & 100 & 56.3\% \\ 
3.4mm (85\%)   & 0.4mm & 56  & 56.0\% \\ 
  & 0.5mm & 36  & 56.3\% \\ 
  & 0.6mm & 25  & 56.3\% \\ \hline 
& 0.3mm & 80  & 45.0\% \\ 
\makecell{3.0mm (75\%)\\}   & 0.4mm & 45  & 45.0\% \\ 
  & 0.5mm & 29  & 45.3\% \\ 
  & 0.6mm & 20  & 45.0\% \\ \hline
 & 0.3mm & 59  & 33.2\% \\ 
2.6mm (65\%)  & 0.4mm & 33  & 33.0\% \\ 
  & 0.5mm & 21  & 32.8\% \\ 
  & 0.6mm & 15  & 33.8\% \\ \hline 
\end{tabular} 
\end{table}

\begin{figure*}[t]
    \centering
    \includegraphics[width=0.85\textwidth]{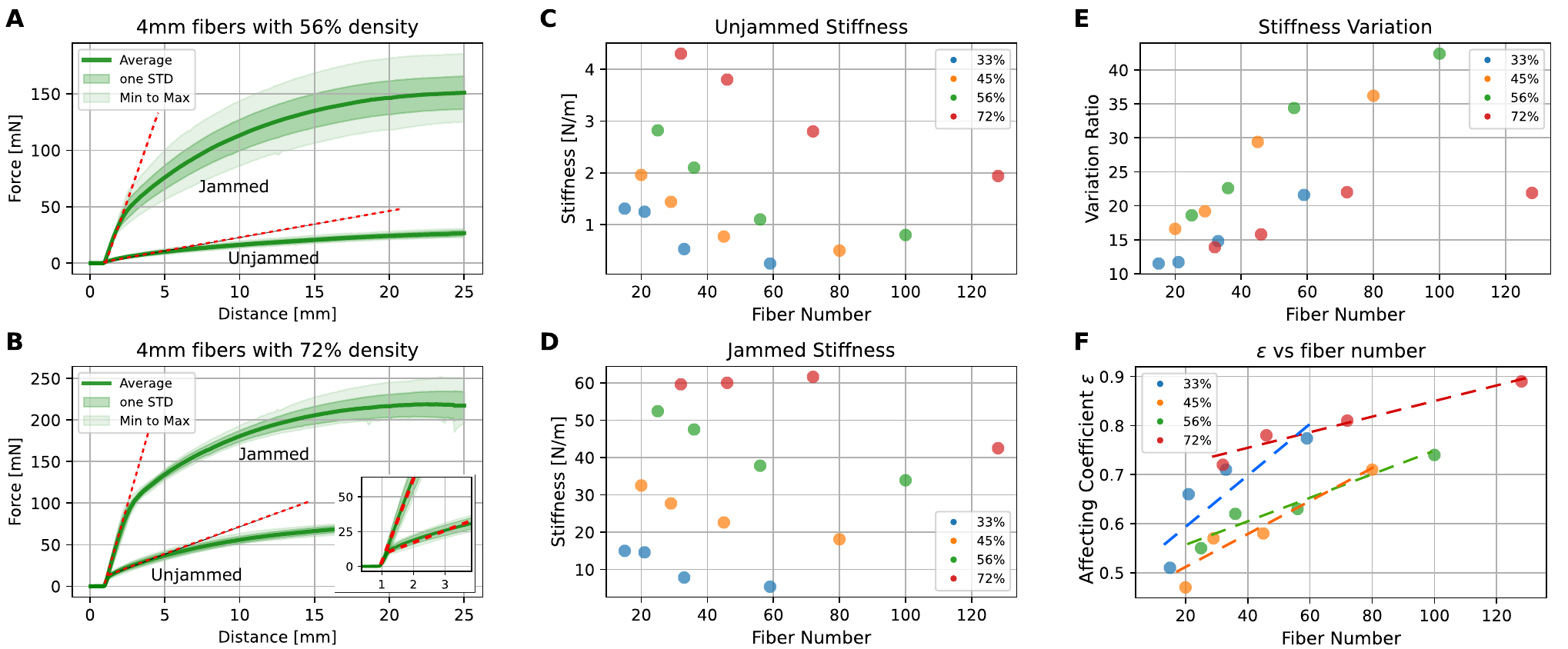}
    \caption{Experimental result: Force vs deflection plots for FJMs with 0.4mm fibers at the density of 56\% \textbf{(A)} and 72\% \textbf{(B)}. \textbf{(C)} The contour plot of unjammed stiffness. \textbf{(D)} The contour plot of jammed stiffness. \textbf{(E)} The contour plot of stiffness variation ratio. \textbf{(F)} The change of coefficient $\epsilon$ respect to fiber number.}
    \label{test result}
\end{figure*}

Fibers are arranged in bundle-type (BT) configuration \cite{brancadoro2020fiber}, with their bottoms fixed using epoxy resin for support. A 0.6mm air hole was left in the center of the support for vacuuming and unvacuuming. Fig.~\ref{Fabricate}B shows the fabrication process with the following steps:
\begin{enumerate}
  \item Select the desired number of fibers and thread them through a 4mm inner diameter silicone tube, which matches the inner diameter of the membrane and is used as a mold for the resin support. 
  \item Thread one 0.6mm fiber through the center of the fiber bundle.
  \item Inject the epoxy resin (5-min Araldite epoxy glue) into the molding tube. Ensure the center fiber is sticking out of the resin during the injection.
  \item After the resin is cured, remove the silicone tube.
  \item Remove the center fiber to create the air channel connecting the outside with the membrane chamber.
\end{enumerate}

In this paper, packing density is defined as the cross-sectional area ratio of the membrane chamber to fibers, which can be represented as: 
\begin{equation}
    \rho=\frac{N (\pi r^2)}{\pi R^2}=\frac{N r^2}{R^2}
\label{density}.
\end{equation}
To establish reasonable intervals between each tested density, we introduced a model to calculate the theoretical maximum number of fibers that can be fitted into a tube in 95\%, 85\%, 75\%, and 65\% of the membrane diameter (4mm) \cite{huang2011global}. 

\subsubsection{Vacuuming Device}
As illustrated in Fig.~\ref{Fabricate}C, the vacuum device comprises a 50ml syringe, silicone tubes, and a digital manometer (EM510 ELECALL). A 3D-printed frame is used to fix the syringe while vacuuming.

\subsection{Stiffness Test}
\subsubsection{Test Setup}
Fig.~\ref{Fabricate}D shows the setup of stiffness tests. To minimize the influences of gravity on the test results, FJM was fixed to the support frame with its tip facing downwards. During the tests, a 1N force gauge (M7-025, Mark-10) was pushing the tip of FJM. A step motor (Zaber LC40B-KM02) travelled the force gauge 25mm forward at a speed of 0.5mm/s.

To validate its repeatability, a minimum of 15 tests were conducted for each FJM configuration in both jammed and unjammed conditions. During the tests on jammed FJM, vacuum pressures were measured at 90-91kpa by the manometer.

\subsubsection{Results and Discussion}
Fig.~\ref{test result}A and~\ref{test result}B show examples of how the reaction force changes with FJM deflection in jammed and unjammed states, consistent with the theoretical study in section II. 
The stiffness of the FJM is illustrated as a straight line in the ``pre-slip" phase for the jammed FJM (red dashed line). The stiffness variation ratio is calculated from the gradient difference between the lines. An exception is the unjammed ``over-fed" FJM  (e.g., FJM at a 72\% density in Fig.~\ref{test result}B), which shows similar stiffness to the jammed state when deflection is minimal due to crowding that prevents fiber slipping. This part of the curve is skipped in unjammed stiffness calculations.

\begin{figure}[t]
    \centering
    \includegraphics[width=0.4\textwidth]{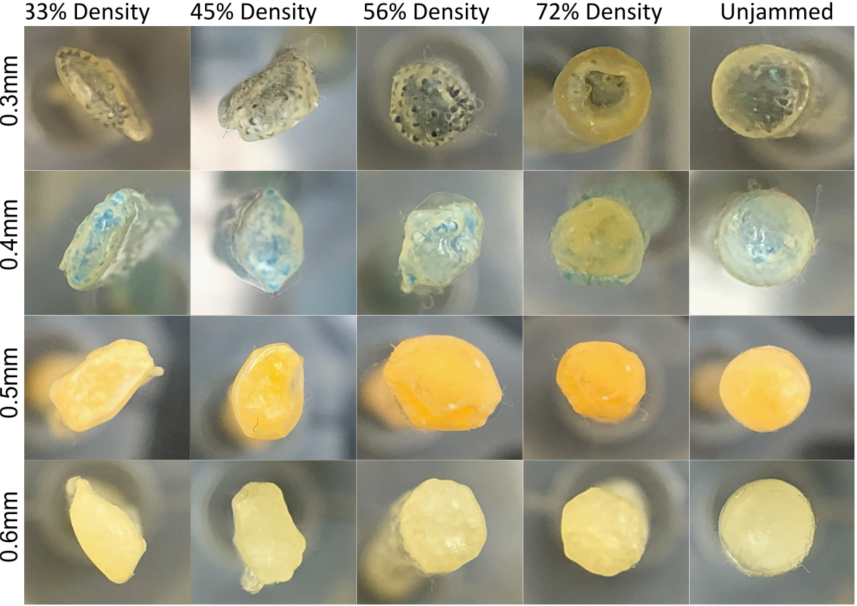}
    \caption{Photos of the cross-section for jammed FJMs showing The cross-section tends to be elliptical as fiber size and density decrease.}
    \label{cross section}
\end{figure}

Fig.~\ref{test result}C-\ref{test result}F summarizes test results for each configuration, including jammed/unjammed stiffness, stiffness variation, and friction coefficient. The following insights are gained:

\begin{enumerate}
    \item \textbf{Tested FJMs achieve up to 4200\% stiffness variation (Fig.~\ref{test result}E)} which is much higher than the larger-size FJMS in previous papers \cite{jadhav2022variable} \cite{brancadoro2020fiber}. One reasonable guess of this result is that we employ more fibers that enhance the difference in stiffness between jammed and unjammed FJMs. However, the allowable deflections are minimal before fibers slippage.
    \item \textbf{The stiffness of unjammed FJMs increases as packing density increases and fiber number decreases (Fig.~\ref{test result}C)}. This is in line with our theoretical conclusion in equation (\ref{I_j}) if we rewrite it as $\frac{\pi}{4}\cdot(N\:r^2)^2/N$ where $N\:r^2$ is proportional to packing density.
    \item \textbf{The stiffness of jammed FJM shows a similar trend as unjammed FJMs (Fig.~\ref{test result}D).} Its change within packing density is in line with equation (\ref{I_j}), where higher density leads to a higher overall diameter $I_j$, and hence a higher stiffness. Theoretically, the jammed stiffness should be independent of the fiber number when the packing density is consistent. However, we see a slow decrease as the fiber number increases due to the unexpected slippage between individual fibers in the ``pre-slip" phase. This finding suggests that we should avoid using fibers that are too small to prevent stiffness loss when jammed.
    \item \textbf{The affecting coefficient $\epsilon$ shows a linear increment with the increasing fiber number (Fig.~\ref{test result}F).} This observation is in line with our theoretical conclusion in equation (\ref{stiffness_ratio}). 
    \item \textbf{The affecting coefficient $\epsilon$ should also be positively correlated with the packing density because it increases the friction force.} Such deduction is evident at the density of 72\% and 56\% but less so at 45\% and 33\%, where the cross-sectional area becomes too elliptical when jammed, causing the equivalent value $R$ to be less than predicted (Fig.~\ref{cross section}). As a result, stiffness variation is smaller than expected, leading to an overestimation of $\epsilon$. 
    \item \textbf{The optimized configuration is achieved by the FJM with 0.4mm fibers at a density of 56\%.} Although its stiffness variation ratio is 3400\%, less than the FJM with 0.3mm fibers at a density of 56\%, its higher jammed stiffness enhances the payload capacity of the FTL robot.
\end{enumerate}
    
\subsection{Application}
In summary, experimental data generally agree with theoretical expectations. An FJM with smaller fibers results in lower jammed and unjammed stiffness but a higher stiffness variation ratio. The optimal packing density for FJM is between 45\% and 56\%, filling 75\% to 85\% of the membrane's space. These findings help us identify the best configuration for designing a MIS robot with FJM.

To further verify the feasibility of the robotic application, we built a prototype to manually demonstrate the FTL motion.
According to the above results, although FJM with 0.3mm fiber has a better stiffness variation ratio, its absolute stiffness is not high enough when jammed. 0.4mm fiber was then selected to balance these two values. 
We chose a 56\% packing density over the highest density of 72\% to avoid excessive stiffness variation and to prevent the cross-sectional area from becoming too elliptical, which would increase resistance during propagation.  Overall, the final prototype (Fig.~\ref{FTL}E) consists of four FJMs with 180mm in length, 4mm in diameter, and 56\% packing density of 0.4mm fibers.


As illustrated in Fig.~\ref{FTL}E and demonstrated in the supplementary video at \href{https://youtu.be/7pI5U0z7kcE}{https://youtu.be/7pI5U0z7kcE}, the prototype can perform an approximately FTL motion with control of the tendon and propagation of FJMs. 
The proposed FJMs demonstrate a balanced stiffness variation between soft and rigid phases. The prototype maintained its shape with jammed FJMs and allowed for flexible steering with unjammed FJMs.
In addition, the actuators currently limit the speed of motion. For example, the vacuuming and unvacuuming motions of an FJM are controlled by a linear actuator, which takes 4 seconds to complete. Such time can be significantly reduced with an appropriate pump. Therefore, the proposed mechanism is capable of higher efficiency by optimizing the actuators.

In summary, the proposed FJMs have significant potential for use in the FTL of robots and could enhance instrument navigation in medical applications, including transanal and transoral surgery.

\section{Conclusion}
In this paper, we presented a theoretical and experimental study on fiber jamming - a potential mechanism of stiffness variation that is suitable for MIS continuum robots to achieve FTL motion. We have deducted the stiffness equation of an FJM relevant to various inputs including fiber size, packing density, etc. We designed and fabricated the FJMs with diameters of 4mm to determine their performance for potential MIS applications. The experimental findings confirmed that both fiber size and packing density have a significant impact on the absolute stiffness and stiffness variation of FJMs, informing the optimal configurations for achieving desired characteristics regardless of the size of the FJMs. 
These findings contribute significantly to enhancing the feasibility of integrating micro-size FJMs into MIS robots while other factors studied in previous research such as fiber material and geometry may not be applicable.


\section{Acknowledgement}
The authors would like to thank Dr. Shuhua Peng for helping with the stiffness test equipment, A/Prof. Jiangtao Xu for advice on the fabrication of FJM, and Prof. Wei Gao for advice on the theoretical study of fiber jamming.

\bibliographystyle{IEEEtran}
\bibliography{citations}

\vspace{12pt}

\end{document}